\documentclass{article}

\usepackage[final]{neurips_2022}

\usepackage{natbib}
\setcitestyle{authoryear,open={(},close={)}}

\usepackage[utf8]{inputenc} 
\usepackage[T1]{fontenc}    
\usepackage[hidelinks]{hyperref}       
\usepackage{url}            
\usepackage{booktabs}       
\usepackage{amsfonts}       
\usepackage{nicefrac}       
\usepackage{microtype}      
\usepackage[table]{xcolor}
\usepackage[pdftex]{graphicx}
\usepackage{bm}
\usepackage{amsmath}
\usepackage{amsfonts}
\usepackage{color}
\usepackage[makeroom]{cancel}
\usepackage{cancel}
\usepackage{tabu}
\usepackage[normalem]{ulem} 
\usepackage{framed}
\usepackage{subcaption}
\usepackage{wrapfig}
\usepackage{amssymb}
\usepackage{algorithm}
\usepackage{algorithmic}
\usepackage{pifont}
\usepackage{multirow}
\usepackage{booktabs}
\usepackage[flushleft]{threeparttable}
\usepackage{float}
\usepackage[capitalize]{cleveref}

\def\0{{\mathbf 0}}

\newcommand{\Zcal}{\mathcal{Z}}

\newcommand{\Acal}{\mathcal{A}}

\newcommand{\Ncal}{\mathcal{N}}

\newcommand{\Lcal}{\mathcal{L}}
\newcommand{\Rcal}{\mathcal{R}}

\newcommand{\xc}{\bm{x}}
\newcommand{\yc}{\bm{y}}
\newcommand{\zc}{\bm{z}}

\newcommand{\Ebb}{\mathbb{E}}

\newcommand{\maxf}[1]{{\cellcolor[gray]{0.87}} #1}

\title{Revisiting Active Sets for Gaussian Process Decoders}

\author{%
  Pablo Moreno-Mu\~{n}oz$^*$ ~~~~~ Cilie W. Feldager\thanks{Equal contribution.} ~~~~~~ Søren Hauberg\\
  Section for Cognitive Systems \\
  Technical University of Denmark (DTU)\\
  \texttt{\{pabmo,cife,sohau\}@dtu.dk} \\
}

\begin{document}

\maketitle
\everypar{\looseness=-1}

\begin{abstract}
Decoders built on Gaussian processes (GPs) are enticing due to the marginalisation over the non-linear function space. Such models (also known as GP-LVMs) are often expensive and notoriously difficult to train in practice, but can be scaled using variational inference and inducing points. In this paper, we revisit active set approximations. We develop a new stochastic estimate of the log-marginal likelihood based on recently discovered links to cross-validation, and we propose a computationally efficient approximation thereof. We demonstrate that the resulting stochastic active sets (SAS) approximation significantly improves the robustness of GP decoder training, while reducing computational cost. The SAS-GP obtains more structure in the latent space, scales to many datapoints, and learns better representations than variational autoencoders, which is rarely the case for GP decoders.\looseness=-1
\end{abstract}

\section{Introduction} \label{sec:intro}
Generative models can be viewed as regression models from unknown inputs. That is, we assume $\xc = f(\zc)$, where $f$ is an unknown mapping from latent variables $\zc$ to observations $\xc$. Given the inherent difficulty of this task, it is perhaps sensible to marginalize the unknown mapping $f$ to avoid the brittleness of point estimates. This is the driving idea in the \emph{Gaussian process latent variable} (GP-LVM) \citep{lawrence2005probabilistic}, which places a Gaussian process (GP) prior over the unknown mapping $f$ and marginalizes accordingly. This contrasts contemporary generative models that predominantly operate with point-estimates of $f$ \citep{kingma2013auto, sohl2015deep}. While conceptually elegant, the GP-LVM is, however, notoriously difficult to train and the conceptual benefits are often not realized in practice.

Exact inference involves computing the marginal likelihood, but (like other GP methods) its cubic complexity in the number of observations $\mathcal{O}(N^3)$ limits the scalability of the GP-LVM.
However, the idea of marginalizing the decoder is sufficiently attractive to motivate the development of scalable and reliable training techniques: Its Bayesian formulation \citep{titsias2010bayesian} variationally integrates out the latent variables $\zc$ to obtain an evidence lower bound.
Using auxiliary inducing variables, \citet{snelson2006sparse} expanded GP regression, which is also applicable in the unsupervised learning setting (i.e. the GP-LVM).

However, considering inducing variables here involves dangers. First, the convergence of inducing points is well-studied in the supervised GP scenario, where \emph{inputs are fixed}, but it differs from the unsupervised case where the \emph{inputs are estimated}. \citet{bauer2016understanding} notes that even in the supervised setting, inducing points are ``\emph{not completely trivial to optimise, and often tricks [...] are required}'', and we hypothesize that this is further complicated in the unsupervised setting where we optimize both latent and inducing variables while they interact.

\textbf{In this paper}, we revisit \emph{active sets} for scaling GP decoders, a sparse approximation predominantly used before the seminal work of \citet*{snelson2006sparse}. From a practical viewpoint, active sets are \underline{fixed} inducing variables that belong to the training dataset. We make links between such active sets and cross-validation, allowing us to lean on a recent result from \citet{fong2020marginal} which, in turn, links cross validation and the log-marginal likelihood. We show how these links allow for a stochastic estimate of the log-marginal likelihood, and that active sets can be seen as a computationally efficient approximation of this. Practically, this amounts to repeatedly and randomly sampling active sets rather than trying to find the optimal active set.
We denote this framework as \emph{stochastic active sets (SAS)}.
We demonstrate that SAS consistently results in significantly better-fitted GP decoders over models trained using inducing points.

\textbf{Historical remarks.}~ Methods based on subsets of data diminish the computational demand and were first introduced in a GP context in the foundational work on sparse approximations by \citet{smola2001sparse}. Back then, \citet{quinonero2005unifying} had already pointed out the main difficulties behind the optimal selection of subsets:
\begin{quotation}
	\noindent
	\hspace{-2mm}``\emph{Traditionally, sparse models have very often been built upon a carefully chosen subset of the training inputs. [...] In sparse Gaussian processes it has also been suggested to select the inducing inputs $\mathbf{X}_{\mathbf{u}}$ from among the training inputs. Since this involves a prohibitive combinatorial optimization, greedy optimization approaches have been suggested [...]. Recently, \citet{snelson2006sparse} have proposed to relax the constraint that the inducing variables must be a subset of training/test cases, turning the discrete selection problem into one of continuous optimization.}''\looseness=-1
\end{quotation}
This explains how inducing variables reshaped the Gaussian process community, effectively banishing other subset-based methods. Our work builds on the advances made in stochastic optimization in the time since active sets were left behind. We show a third way over those considered by \citet{quinonero2005unifying}: instead of \emph{optimizing} the active set, we \emph{average} with respect to it. This simplifies matters notably and makes them more robust.

To justify our approach, we establish a link between active sets and cross validation (CV). The latter has a long history for model selection in GPs, dating at least to the seminal work of \citet{wahba1990spline}.
For probabilistic models, \citet{williams2006gaussian} point to the utility of CV variants within negative log-probabilities. Building on results from \citet{fong2020marginal} linking CV and log-marginal likelihoods, we argue that, for GP-LVMs, active sets combine more gracefully with stochastic optimization. The remainder of this paper elaborates on this viewpoint and demonstrates it empirically.\looseness=-1

\section{Gaussian Process Decoders}
\label{2_bayesian_gp}

The Gaussian process latent variable model (GP-LVM) \citep{lawrence2005probabilistic}
defines a \emph{decoder} which is a non-linear mapping\footnote{We use $\{\xc, \zc\}$ to denote observations and latent variables respectively, since we do not consider inducing variables. Notice that \citet{lawrence2005probabilistic} use the notation $\{\yc, \xc\}$.} $\xc = f(\zc)$ from the latent space $\mathcal{Z}\in \mathbb{R}^Q$ to observation space $\mathcal{X}\in \mathbb{R}^D$. The prior on this map is a Gaussian process (GP) so it is drawn like $f \sim \mathcal{GP}(0, k_{\bm{\theta}}(\cdot, \cdot))$, where $k_{\bm{\theta}}$ is the covariance function or kernel and $\bm{K}_{NN}$ denotes the evaluated kernel function so the $i,j$th element of $\bm{K}_{NN}$ equals $k_{\bm{\theta}}(z_i, z_j)$. For clarity, we omit the dependence on covariance function parameters, $\bm{\theta}$.

The original version of the GP-LVM starts from one-dimensional observations $\xc = \{\xc_n\}^N_{n=1}$ and latent variables $\zc = \{\zc_n\}^N_{n=1}$, and factorises the joint distribution of the model as $p(\xc, f| \zc) = p(\xc| f, \zc) p(f|\zc)$. Here the conditional distributions correspond to the likelihood model and the prior
\begin{equation}
     p(\xc| f, \zc) = \prod^N_{n=1}\Ncal(\xc_n|f(\zc_n), \sigma^2), \hspace{2cm} p(f|\zc) = \Ncal(f(\zc)|0, \bm{K}_{NN}).
     \label{eq:likelihood_GPLVM}
\end{equation}%
When the data dimensionality is $D\!>\!1$, the model factorises across dimensions, and we have different mappings $f$ per $d^{\mathrm{th}}$ feature. One of the principal assumptions of the GP-LVM is that the prior $p(f)$ regularises the smoothness of all mappings equally, so we only consider one \emph{kernel}. This assumption can be relaxed if needed, but at increased computational cost and with more learnable hyperparameters.

\textbf{Mapping marginalization.}~ A GP prior over the non-linear decoder $f$ allows for marginalisation of $f$ to obtain a closed-form expression of the marginal likelihood of the GP-LVM
\begin{equation*}
p(\xc|\zc) = \int p(\xc| f, \zc)p(f|\zc) \mathrm{d}f = \Ncal(\xc|0, \bm{K}_{NN} + \sigma^2\mathbb{I}).
\end{equation*}
On a $\log$-scale, this gives the following objective function \citep{lawrence2004gaussian}, which can be optimized w.r.t.\@ both hyperparameters $\bm{\theta}$ and latent variables $\zc$
\begin{equation}
  \Lcal = -\frac{DN}{2}\log 2\pi - \frac{D}{2}\log |\bm{K}_{NN} + \sigma^2\mathbb{I}| - \frac{1}{2}\text{tr}(\left(\bm{K}_{NN} + \sigma^2\mathbb{I}\right)^{-1}\xc\xc^\top).
  \label{eq:log_marginal}
\end{equation}
The difficulties of training the GP-LVM using this objective function are evident above, as the evaluation cost grows cubically with the number of observations $N$. Furthermore, notice that observations $\xc$ are no longer independent (in contrast with Eq.~\ref{eq:likelihood_GPLVM}) once $f$ is integrated out. The log-marginal likelihood will be the starting point for our approach in Sec.~\ref{sec:sas_3}.

\textbf{Bayesian extension.} In the seminal works of \citet{lawrence2004gaussian,lawrence2005probabilistic}, the GP-LVM is first derived as a non-linear extension of probabilistic principal component analysis (\textsc{ppca}) \citep{tipping1999probabilistic}. Considering the \emph{isotropic} prior on the latent variables $\zc$, such that $p(\zc_n) = \Ncal(\zc_n|0, \mathbb{I})~~\forall \zc_n\in \zc $, the general idea is to optimize them rather than introducing marginalization. From a \emph{full} probabilistic perspective, one could also be interested in the posterior distribution over $\zc$, which leads to using Bayesian inference for the GP-LVM approach. This is the driving idea of \citet{titsias2010bayesian}, where variational methods are introduced. In particular, latent variables are not easy to marginalize, mainly due to their presence in the kernel mappings, so a lower-bound on the log-marginal likelihood $\log p(\xc) = \log \int p(\xc|\zc)p(\zc)\mathrm{d}\zc$ of the model is derived.

So far, the Bayesian GP-LVM model \citep{titsias2010bayesian} has been considered as the standard methodology to apply GPs to large unsupervised datasets with $10^4$$-$$10^6$ observations, e.g.\@ in regression \citep{bui2015stochastic}, classification \citep{gal2015latent} and representation learning \citep{martens2019decomposing} tasks. 

\section{Stochastic Active Sets}
\label{sec:sas_3}

Our key objective is a computationally efficient estimate of the log-marginal likelihood in Eq.~\ref{eq:log_marginal}, as this is known to be a good measure of generalization performance \citep{rasmussen2000occam,germain2016pac}. Another popular measure of generalization performance is \emph{cross validation} (CV) \citep{geisser1979predictive,vehtari2002bayesian}, which is arguably mostly used outside the realm of Bayesian models. Recently, \citet{fong2020marginal} linked these two measures, effectively showing that the marginal likelihood is equivalent to the average over exhaustive leave-$R$-out CV scores. In particular, the average is w.r.t.\@ the size of the hold-out set. More precisely, let
\begin{equation}
    \mathcal{S}_{\text{CV}}(\xc|R) = \frac{1}{\mathcal{C}}\sum^{\mathcal{C}}_{p=1}\frac{1}{R}\sum_{n\in \Rcal_p} \log p(\xc_{n}|\xc_{\Acal_p},\zc) = \frac{1}{R}\Ebb_{\Acal_p}\left[\sum_{n\in \Rcal_p} \log p(\xc_{n}|\xc_{\Acal_p},\zc)\right],
    \label{eq:cv_score}
\end{equation}
denote the leave-$R$-out CV using log-predictive scoring functions $\log p(\xc_{n}|\xc_{\Acal_p}, \zc)$. Here $\Acal_p$ denotes the \emph{active set} indices of the training data, such that $\Acal_p \subset \{1,2,\dots, N\}$ and $\Rcal_p = \{1,2,\dots, N\} \setminus \Acal_p$ are the remaining hold-out samples. The subscript $p\in \mathcal{C}$ denotes the permutation, and we average over all $\mathcal{C} = \binom{N}{R}$ possible hold-out sets. We use use $R$ to indicate the size of the hold-out set $\Rcal_p$ {\color{black}($R \equiv |\mathcal{R}_p|$)} and let $A\equiv|\Acal_{p} |= N-R$. If we average $\mathcal{S}_{\text{CV}}(\xc|R)$ over all possible sizes of the hold-out set, then \citet{fong2020marginal} has shown that we recover the exact log-marginal likelihood,
\begin{equation}
    \log p(\xc|\zc) = \sum^{N}_{r = 1}\mathcal{S}_{\text{CV}}(\xc|r) = \mathcal{S}_{\text{CCV}}(\xc|R) + \mathcal{S}_{\text{PCV}}(\xc|R).
    \label{eq:cv_marginal}
\end{equation}

Here, $\mathcal{S}_{\text{PCV}}(\xc|R) = \Ebb_\Acal[\log p(\xc_\Acal|\zc_\Acal)]$ and $\mathcal{S}_{\text{CCV}}(\xc|R) = \sum^{R}_{r = 1}\mathcal{S}_{\text{CV}}(\xc|r)$ is the cumulative CV score, which reduces to a sum of expectations over the predictive factors.\footnote{We drop the permutation subscript $p$ in $\Acal$ and $\Rcal$ to avoid cluttered notation.} Further details on $\mathcal{S}_{\text{PCV}}$ and $\mathcal{S}_{\text{CCV}}$ can be found in the Appendix. \citet{fong2020marginal} used Eq.~\ref{eq:cv_marginal} to argue in favor of using the marginal likelihood over cross-validation for model selection. 
\begin{figure}[t!]
\centering
\includegraphics[width=\textwidth]{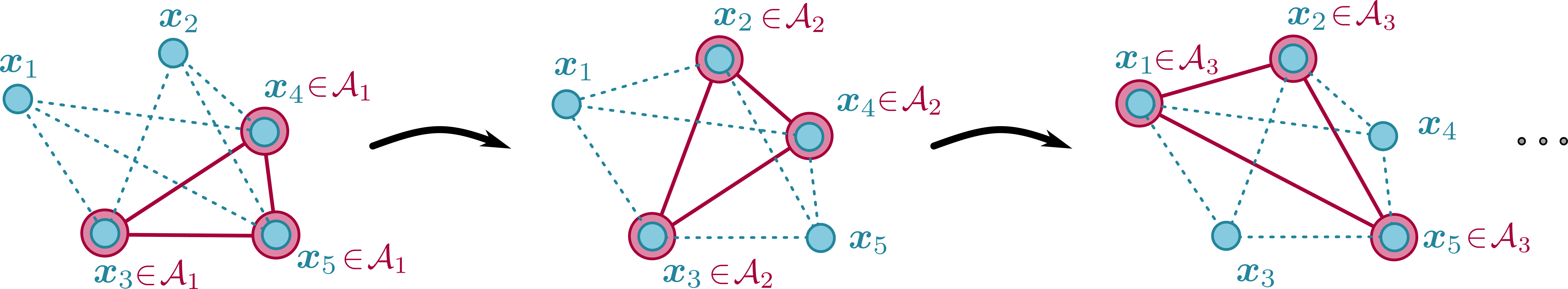}
\caption{Schematic graphical model of the correlation structure given different permutations $p$ of the active set $\Acal_1, \Acal_2, \dots, \Acal_\mathcal{C}$ for five observations $\{\xc_1,\dots,\xc_5\}$. Thick red lines indicate that we build \emph{full} covariance densities between observations included in $\xc_{\Acal}$. Only-blue variables are considered conditionally independent among them w.r.t.\ to the red colored ones. Dashed lines indicate the conditional probability factors $p(\xc_n|\xc_\Acal, \zc)$ that we can \emph{easily} compute.}
\label{fig:active_set}
\end{figure}

\textbf{Stochastic approximation.~~} We take a slightly different view than \citet{fong2020marginal} and argue that Eq.~\ref{eq:cv_marginal} can be the grounds for an efficient stochastic gradient \citep{robbins1951stochastic} of the log-marginal likelihood suitable for training. In the context of GPs, we have that conditional probabilities for $\mathcal{S}_{\text{PCV}}(\xc|R)$ and  $\mathcal{S}_{\text{CCV}}(\xc|R)$ in Eq.~\ref{eq:cv_marginal} are
\begin{equation}
    p(\xc_\Acal|\zc_\Acal) = \Ncal(\xc_\Acal|0, \bm{K}_{\Acal\Acal} + \sigma^{2}_n\mathbb{I}), \hspace{1cm}
    p(\xc_{n}|\xc_{\Acal},\zc) = \Ncal(\xc_n|\bm{m}_{n|\Acal}, \bm{c}_{n|\Acal}),
    \label{eq:conditional}
\end{equation}
where we used Eq.~2.22 from \citet{williams2006gaussian} to obtain
\begin{align*}
    \bm{m}_{n|\Acal} &= \bm{K}_{n\Acal}(\bm{K}_{\Acal \Acal} + \sigma^2_n\mathbb{I})^{-1}\xc_\Acal,\\
    \bm{c}_{n|\Acal} &= \bm{K}_{nn} + \sigma^2_n\mathbb{I} - \bm{K}_{n\Acal}(\bm{K}_{\Acal\Acal} + \sigma^2_n\mathbb{I})^{-1}\bm{K}^\top_{n\Acal},
    \label{eq:conditional_cov}
\end{align*}
and $\bm{K}_{\Acal\Acal} \in \mathbb{R}^{A\times A}$ has entries $k(\zc_{i},\zc_{j})$ with
$\zc_{i},\zc_{j} \in \zc_{\Acal}$. The computational cost of $\mathcal{S}_{\text{PCV}}(\xc|R)$ is $\mathcal{O}(A^3)$, while $\log p(\xc_{n}|\xc_{\Acal},\zc)$ can reuse the matrix inversion $\bm{K}^{-1}_{\Acal\Acal}$ to only have an additional linear cost. Clearly, we can obtain an \emph{unbiased} stochastic estimate of the log-marginal likelihood by \emph{first} uniformly sampling $R$, \emph{second} making a random split permutation into train and hold-out data $\Rcal$ and \emph{finally} by evaluating 
\begin{equation}
      \log p(\xc|\zc) \approx \sum_{n\in \Rcal} \log p(\xc_{n}|\xc_{\Acal},\zc)
     +  \log p(\xc_{\Acal}|\zc_{\Acal}),
     \label{eq:stochastic_approx}
\end{equation}
where we remark that the summation can be mini-batched. This approach is equivalent to the decomposition $\log p(\xc|\zc) = \log p(\xc_\Rcal|\xc_\Acal,\zc) + \log p(\xc_\Acal|\zc_\Acal)$ and it also assumes conditional independence among observations $\xc_n$ for $n\!\in\!\Rcal$. This is similar to the standard \emph{active set} approximation \citep{seeger2003fast} previously discussed, and we may think of $\Acal$ as a \emph{stochastic active set (SAS)}.

However, the estimate of $\log p(\xc|\zc)$ still has the same computational complexity $\mathcal{O}(N^3)$ as the usual deterministic approach, since we may sample $A\!=\!N$ in the worst case. Instead, we propose to make a stochastic approximation where we choose the size of the active set deterministically through a user-specified parameter, such that the computational cost reduces to $\mathcal{O}(A^3)$, as in sparse approximations based on inducing points. This does not need to be unbiased; empirically we find that most often the approximation behaves as a lower bound to the true marginal likelihood, and that, in all instances, it is a rather close approximation. We include a longer discussion on this point in the Appendix and the training methodology using SAS is in Alg.~\ref{alg:sas}.

\subsection{Extension for Bayesian GP decoders}

We next seek to extend the previous SAS approach to the Bayesian GP-LVM, where we aim to obtain the posterior distribution $p(\zc|\xc)$. From this perspective, we are interested in marginalising latent variables $\zc$ to obtain the marginal likelihood $p(\xc)$ of the model.\footnote{Notice that the probabilistic objective function changes between standard and Bayesian versions of the GP-LVM. In the former case, we usually look for $p(\xc|\zc)$ as the marginal likelihood w.r.t.\ the mapping $f$. This is the one usually considered in supervised GP tasks. In the latter, we refer to $p(\xc)$ as the marginal likelihood of the model, since $\zc$ are also integrated out.} However, this integration is not possible, as latent variables appear non-linearly in the kernel function \citep{titsias2010bayesian}. Alternatively, we consider the variational inference scheme, where an auxiliary distribution $q(\zc)$ is introduced into the formulation. Therefore, we are able to build the evidence lower bound (ELBO) of the model using Jensen's inequality as
\begin{equation}
    \log p(\xc) \geq \int q(\zc)\log p(\xc|\zc)p(\zc)\mathrm{d}\zc = \mathbb{E}_{q(\zc)}\left[ \log p(\xc|\zc) \right] - \text{KL}\left[q(\zc)||p(\zc)\right],
    \label{eq:elbo}
\end{equation}
which is equivalent to the one obtained by \citet[Eq.~8]{titsias2010bayesian}. At this point, the computational cost is $\mathcal{O}(N^3)$, since the ELBO requires evaluating $\log p(\xc|\zc)$, where we invert $\bm{K}_{NN}$. 
The expectation in Eq.~\ref{eq:elbo} can also benefit from a stochastic SAS approximation, just as with inducing points \citep{hensman2013gaussian}. Thus, the lower bound can be approximated as
\begin{equation}
\mathcal{L}_{\text{ELBO}} \approx \sum_{n\in\Rcal}\mathbb{E}_{q(\zc_n)}\left[\log p(\xc_n|\xc_{\Acal}, \zc) \right] + \mathbb{E}_{q(\zc_\Acal)}\left[\log p(\xc_{\Acal}|\zc_{\Acal})\right] - \sum^{N}_{n=1}\text{KL}[q(\zc_n)||p(\zc_n)],
\label{eq:metaelbo}
\end{equation}
where we consider \emph{mean-field} \textsc{vi} to factorize the distribution $q(\zc)$. The Bayesian GP-LVM shares high-level similarities with other generative models that marginalize the latent variable according to a simple prior \citep{rezende2015variational}. The proposed SAS approximation \eqref{eq:metaelbo} scales similarly to mini-batched inducing point approximations, but we will later see that SAS behaves notably better in practice. Algorithmically, the approach is simple, and the summary code is provided in Alg.~\ref{alg:bayesian_sas}.

\begin{minipage}{0.48\textwidth}
\begin{algorithm}[H]
    \centering
    \caption{SAS for GP decoders}
    \label{alg:sas}
    \begin{algorithmic}[1]
        \STATE {\textbf{Input:}} Observed data $\xc$
        \STATE {\bfseries Parameters:} Initialize $\bm{\theta}$, $\bm{\phi}$ \hfill \textcolor{gray}{\small{// $\bm{\theta},\zc$ if \textsc{na}}}
        \FOR{$e$ \textbf{in} epochs}
        \FOR{$b$ \textbf{in} batches}
        \STATE {Sample $\xc_{\text{batch}} \sim \xc$} 
        \STATE {$\xc_\Rcal, \xc_\Acal$ $\leftarrow$ \texttt{random\_split}($\xc_{\text{batch}}$)}
        \IF{amortized}
        \STATE{Get $\{\zc_\Rcal, \zc_\Acal\} \leftarrow g(\xc_\Rcal, \xc_\Acal|\bm{\phi})$}
        \ENDIF
        \STATE{Compute $\bm{K}^{-1}_{\Acal\Acal}$ \hfill \textcolor{gray}{\small{// via Cholesky}}} 
        \STATE {Evaluate $\log p(\xc_\Acal|\zc_\Acal)$}
        \STATE {Evaluate $\log p(\xc_{n}|\xc_\Acal, \zc)$,~ $\forall \xc_n\!\in\!\xc_\Rcal$}
        \STATE {Evaluate Eq.~\ref{eq:stochastic_approx}}
        \STATE {\textbf{do} Adam($\bm{\theta}, \bm{\phi}$) step for $\Lcal$}
        \ENDFOR
        \ENDFOR
    \end{algorithmic}
    {\raggedright\textcolor{gray}{\small{\textsc{na}: Non-amortized.}}\par}
\end{algorithm}
\end{minipage}
\hfill
\begin{minipage}{0.48\textwidth}
\begin{algorithm}[H]
    \centering
    \caption{SAS for Bayesian GP decoders}
    \label{alg:bayesian_sas}
    \begin{algorithmic}[1]
        \STATE {\textbf{Input:}} Observed data $\xc$
        \STATE {\bfseries Parameters:} Initialize $\bm{\theta}$, $\bm{\phi}$ \hfill \textcolor{gray}{\small{// $\bm{\theta},\mu, \sigma$ if \textsc{na}}}
        \FOR{$e$ \textbf{in} epochs}
        \FOR{$b$ \textbf{in} batches}
        \STATE {Sample $\xc_{\text{batch}} \sim \xc$} 
        \STATE {$\xc_\Rcal, \xc_\Acal$ $\leftarrow$ \texttt{random\_split}($\xc_{\text{batch}}$)}
        \IF{amortized}
        \STATE{Get $\mu_{\zc} \leftarrow g_{\mu}(\xc_\Rcal, \xc_\Acal|\bm{\phi})$}
        \STATE{Get $\sigma_{\zc} \leftarrow g_{\sigma}(\xc_\Rcal, \xc_\Acal|\bm{\phi})$}
        \ENDIF
        \STATE{Sample $\{\zc_{\Rcal}, \zc_{\Acal}\} \sim q(\mu_{\zc}, \sigma_{\zc})$ \hfill \textcolor{gray}{\small{// \textsc{rt}}}}
        \STATE{Compute $\bm{K}^{-1}_{\Acal\Acal}$ \hfill \textcolor{gray}{\small{// via Cholesky}}}
        \STATE {Evaluate $\Lcal$ in Eq.~\ref{eq:metaelbo}}
        \STATE {\textbf{do} Adam($\bm{\theta}, \bm{\phi}$) step for $\Lcal_{\text{ELBO}}$}
   \ENDFOR
   \ENDFOR
    \end{algorithmic}
    {\raggedright\textcolor{gray}{\small{\textsc{na}: Non-amortized, \textsc{rt}: Reparametrization trick.}}\par}
\end{algorithm}
\end{minipage}
\subsection{The Role of Amortization}
\label{sec:amortization}
Early after the emergence of the seminal GP-LVM \citep{lawrence2005probabilistic}, the lengthy optimization required obtaining a result in which all latent representations $\zc_n$ became noticeable. An additional consideration is that, while most approaches to non-linear low-dimensionality reduction focus on preserving similarities, the GPLVM does the opposite. This property was initially discussed by \citet{lawrence2006local}, since in some sense the GP-LVM is \emph{dissimilarity preserving}, such that different observations will generally be represented far away from each other. In practice, we are often more interested in embeddings that reflect the \emph{true} distance between the observed objects in their representations, particularly those that are close together. This observation inspired \emph{back constraints} for locality preservation \citep{lawrence2006local}, which enforces latent variables $\zc$ to be an explicit function of observations $\zc = g(\xc|\bm{\phi})$ parameterized by $\bm{\phi}$. This is similar to the stochastic \emph{encoder} applied in variational autoencoders \citep{kingma2013auto, rezende2015variational}. This idea was also extended to the \textsc{vi} framework in GP-LVMs by \citet{bui2015stochastic} using a recognition model, e.g.\@ $q(\zc_n)\!=\! \Ncal(\zc_n|g_{\mu}(\xc_n|\bm{\phi}),g_{\sigma}(\xc_n|\bm{\phi}))$ and more recently, to accelerate hyperparameter learning in GPs with hierarchical attention networks \citep{liu2020task}. \looseness=-1

In our context, we assume the mappings to be neural networks (NNs) like \citet{bui2015stochastic}, referring to them as \emph{amortization} networks. We find such networks accelerate learning very nicely when used in conjunction with SAS. It is also worth noting that amortization has been empirically shown to improve generalization performance \citep{shu2018amortized}.

\section{Related Work}

Marginal likelihood approximations were used in GPs \citep{smola2001sparse, csato2002sparse} before the apparition of pseudo-inputs \citep{snelson2006sparse} and the associated variational inference framework \citep{titsias2009variational}. In the former case, stochastic approximations to the ELBO were first presented by \citet{hensman2013gaussian,hensman2015scalable}, in line with the Bayesian counterpart of SAS. In terms of active sets, \citet{seeger2003fast} empirically observed that the approximation was generally stable for optimisation, even if the size of $\Acal$ was a small fraction of the training size only. However, they also observed that random selection of active sets led to non-smooth fluctuations, making it difficult to converge through exact gradient ascent. Particularly, the issue with re-selecting of $\Acal$ motivated \citet{snelson2006sparse} as a \emph{smoother} optimization alternative, and we find that SAS also circumvents this problem via stochastic estimates as shown in Sec.~\ref{sec:experiments}.

The connection between cross-validation and GPs was already described in \citet{williams2006gaussian} as an alternative for model selection. However, the equivalence between \emph{exhaustive} CV and the log-marginal likelihood provided in \citet{fong2020marginal} provides a novel perspective that we exploit. Additionally, the notion of \emph{back constraints} has recently been considered in \citet{lalchand22a} for GP-LVMs with inducing points, where a doubly stochastic formulation is used. More recently, amortization networks have been used to drastically reduce the number of inducing points in supervised GPs.

\section{Experiments}
\label{sec:experiments}

In this section, we evaluate the performance of the SAS approach for stochastic learning of GP decoders, both the deterministic GP-LVM \citep{lawrence2004gaussian} and its Bayesian counterpart \citep{titsias2010bayesian}. For this purpose, we consider three different datasets: \textsc{mnist} \citep{lecun1998gradient}, \textsc{fashion mnist} \citep{xiao2017fashion} and \textsc{cifar-10} \citep{krizhevsky2009learning}.
For a \emph{fair} comparison of our model with baseline methods, we use the same amortization across models, namely a neural network {\color{black}(three linear layers ReLU activation functions)\footnote{Please see the supplementary material for more details.}} for all models in all experiments and all GPs use an quadratic exponentiated kernel. In all experiments, the learning rates are set in the range $[10^{-4},10^{-2}]$, the maximum number of epochs considered is $300$ and we use the \textsc{adam} optimizer \citep{kingma2014adam}. For SAS experiments, we only consider batch sizes greater than the active set size, as this is a requirement for SAS.

Performance metrics of the SAS-GP decoders are given in terms of the negative log-predictive density (\textsc{nlpd}), root mean-square error (\textsc{rmse}) and mean {\color{black}absolute} error (\textsc{mae}). In all cases, we optimize w.r.t.\ an approximation to the log-marginal likelihood $\log p(\xc|\zc)$ in the deterministic scenario or w.r.t.\ a lower bound on the $\log p(\xc)$ \emph{of the model} in the Bayesian cases.  We also provide \textsc{Pytorch} code that allows for reproducing all experiments and models.\footnote{The code is publicly available in the repository: \url{https://github.com/pmorenoz/SASGP/} including baselines.} We monitor the run-time of convergence as we suspect that rotating active sets (see Fig.\ \ref{fig:active_set}) across the dataset is a fast way to capture the correlation structure of the regression problem.

\begin{figure}[ht!]
	\centering
	\includegraphics[width=\textwidth]{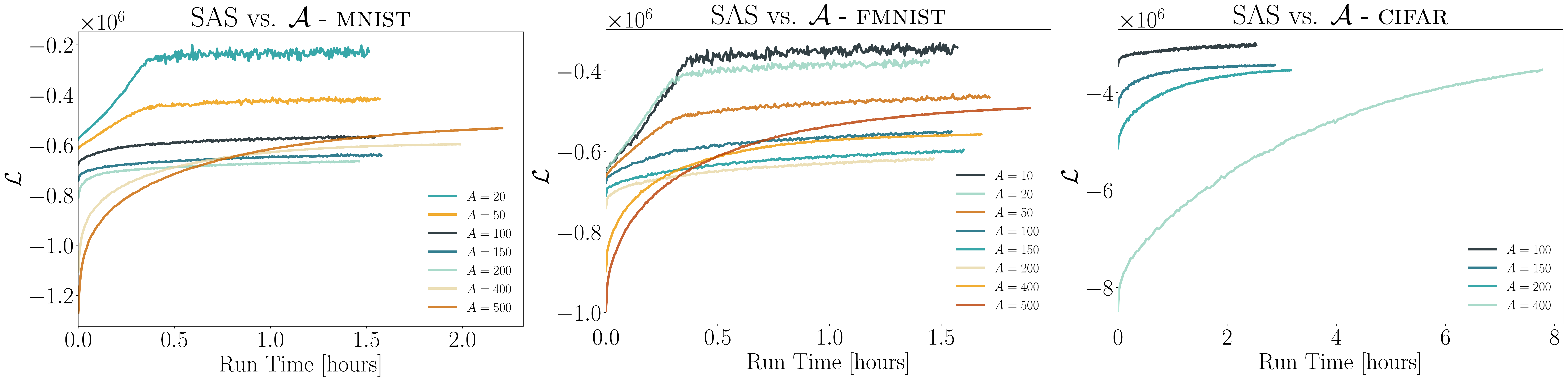}
	\includegraphics[width=\textwidth]{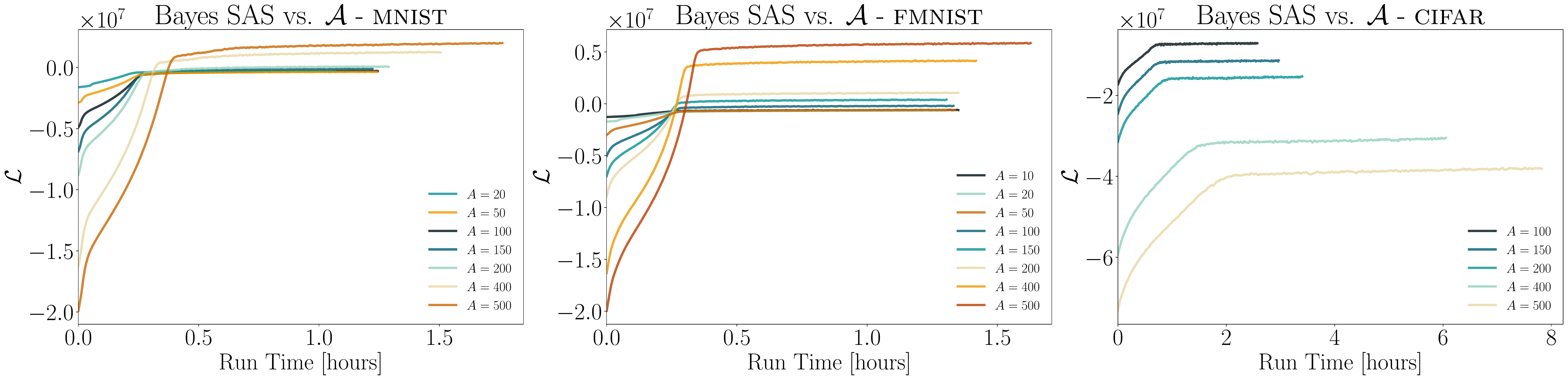}
	\caption{{\color{black}Approximate log-marginal likelihood (\textbf{upper row}) for SAS and \textsc{elbo} curves (\textbf{lower row}) for Bayesian SAS. We fix the batch size in Alg.\ \ref{alg:sas} to be $B\!=\!1024$ and study the convergence for different \emph{active set} sizes $A$. All values in the curves displayed are \emph{per-epoch}.}}
	\label{fig:sas_AS_curves}
\end{figure}

\begin{figure}[ht!]
	\centering
	\includegraphics[width=\textwidth]{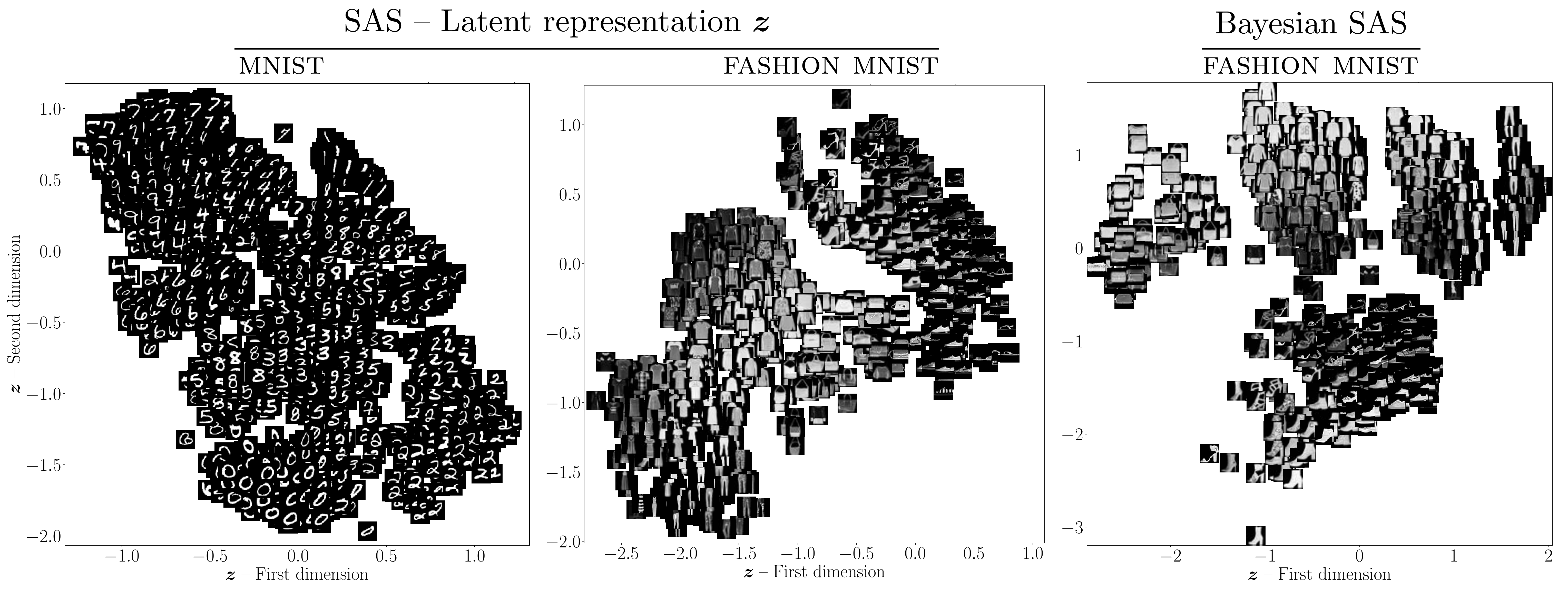}
	\caption{Latent representation in a 2-dimensional space $\mathcal{Z}$ for the 10 \textsc{mnist} and \textsc{fmnist} classes learnt with SAS and Bayesian SAS. Notice that the likelihood model of the GP decoder is controlled by a \textit{vanilla} RBF kernel. The examples have been obtained using an \emph{active set} size $A\!=\!800$.}
\end{figure}
\begin{figure}[ht!]
\centering
\includegraphics[width=1.0\textwidth]{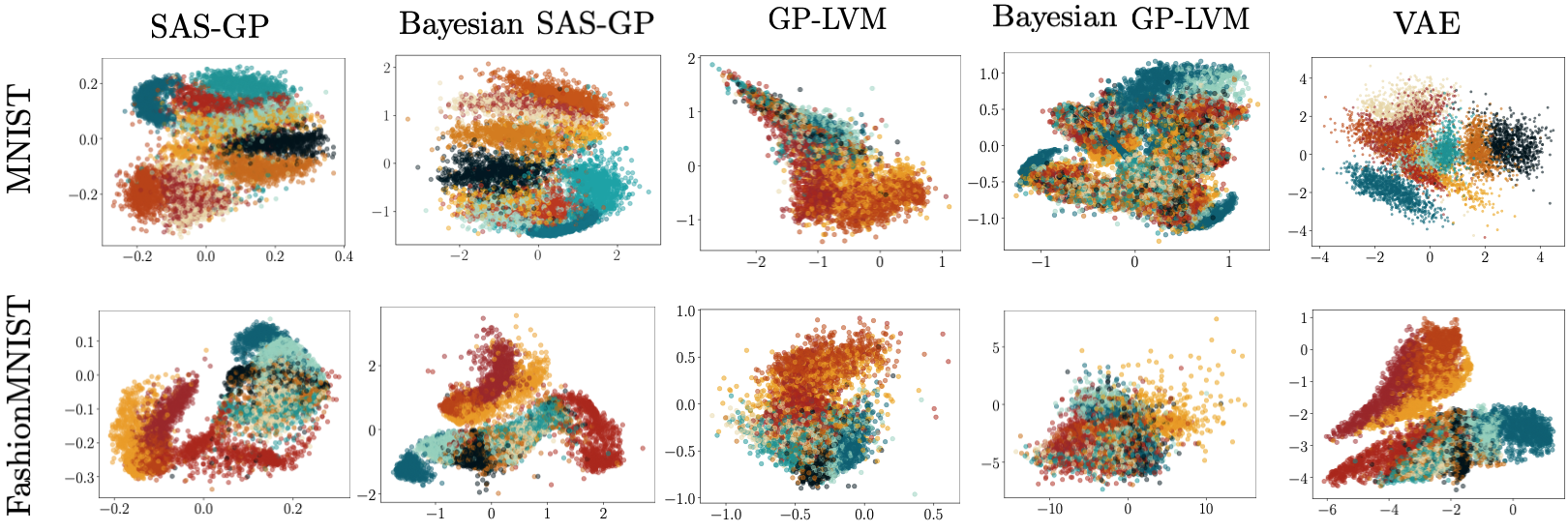}
\caption{Illustration of latent space mappings $\zc_n \in \mathcal{Z}$ on test data for different SAS models and baselines (\textbf{rows}) and different datasets (\textbf{columns}). The models have been trained on \emph{full} \textsc{mnist} and \textsc{fmnist}.}
\label{fig:latent_spaces}
\end{figure}

\subsection{Representation performance}

First, we analysed our SAS approach and the approximate optimization procedure on an \emph{unsupervised} version of \textsc{mnist}, \textsc{fmnist} and \textsc{cifar-10}, where we took the full training corpus for learning two-dimensional latent representations of images. The approximation curves are included in Fig.\ \ref{fig:sas_AS_curves}, where we observe convergence in less that 2h runtime in CPU for most cases. For all experiments, we can observe that for larger active set sizes $A$, the SAS approaches take longer time to complete the $300$ epochs, as the computational cost of inversion is higher.

\textbf{Evaluation with SOTA methods.}~~ We also tested the performance of representation learning in the state-of-the-art methods with and without GPs for unsupervised scenarios. Results are shown in Fig.\ \ref{fig:latent_spaces}. We include a short description of the models considered:
\vspace{-\topsep}
\begin{itemize}
    \itemsep0em 
     \item \textsc{sas-gp decoder ---} It uses stochastic active sets to approximate the log-marginal likelihood $\log p({\xc|\zc})$. The training methodology is described in Alg.\ \ref{alg:sas}.
    \item \textsc{Bayesian sas-gp decoder ---} It uses stochastic active sets to approximate the ELBO in Eq.\ \eqref{eq:metaelbo} on $\log p({\xc})$. See Alg.\  \ref{alg:bayesian_sas}.
    \item \textsc{Bayesian gp-lvm ---}  It is based on the model in \citet{titsias2010bayesian}. Means and variance parameters are generated  by an amortization \textsc{nn} as in \citet{bui2015stochastic}. The model is trained using stochastic variational inference \citep{hensman2013gaussian}.
    \item \textsc{gp-lvm --- } We used the model proposed by \citet{lawrence2005probabilistic} enhanced with an \textsc{nn} encoding to the latent space. The model is trained using maximum likelihood (ML). 
    \item \textsc{Variational autoencoder (vae) ---}  \citep{kingma2013auto} The \textsc{nn} encoder has the \underline{same} architecture as the amortization function used in the SAS-GP models.
\end{itemize}
For the SAS-GP decoder and GP-LVM, a neural network encodes the latent variables. Likewise, for the Bayesian SAS-GP decoder and the Bayesian GP-LVM, two neural networks each encode the latent means and latent variances. We refer to this encoding as \emph{amortization}. All models use a Gaussian likelihood\footnote{Pytorch \citep{paszke2019pytorch} and Pyro \citep{bingham2019pyro} are available on public repositories.}.

\begin{table}[h!]
	\caption{Comparative metrics for SAS and Bayesian SAS on \textsc{mnist}, \textsc{fmnist} and \textsc{cifar-10}.}
	\resizebox{\textwidth}{!}{
		\color{black}\scriptsize
		\begin{tabular}{ccccccc}
			\toprule
			\textsc{model} & & \textcolor{black}{\textsc{sas}} & & & \textcolor{black}{\textsc{bayesian sas}} & \\
			\midrule
			\textsc{active set size}  & \textcolor{black}{$A=100$} & \textcolor{black}{$A=200$} & \textcolor{black}{$A=400$} & \textcolor{black}{$A=100$} & \textcolor{black}{$A=200$} & \textcolor{black}{$A=400$} \\
			\midrule
			\textsc{mnist~/~rmse} $\downarrow$ & $2.55 \pm 0.98$ & $2.47 \pm 0.98$ & $\maxf{2.41 \pm 0.93}$ & $2.16 \pm 0.02$ & $2.08 \pm 0.02$ & $\maxf{1.99 \pm 0.02}$ \\
			\textsc{mnist~/~mae} ~\,$\downarrow$ & $1.61 \pm 0.97$ & $1.55 \pm 0.99$ & $\maxf{1.51 \pm 0.96}$ & $1.11 \pm 0.02$ & $1.04 \pm 0.02$ & $\maxf{0.96 \pm 0.01}$ \\
			\textsc{mnist~/~nlpd} $\downarrow$& $2.99 \pm 1.41$ & $2.92 \pm 1.38$ & $\maxf{2.84 \pm 1.31}$ & $2.33 \pm 0.03$ & $2.26 \pm 0.02$ & $\maxf{2.17 \pm 0.02}$ \\
			&&&&&&\\
			\textsc{fmnist~/~rmse} $\downarrow$& $2.37 \pm 0.95$ & $2.31 \pm 0.94$ & $\maxf{2.25 \pm 0.90}$ & $1.99 \pm 0.17$ & $1.88 \pm 0.20$ & $\maxf{1.85 \pm 0.13}$ \\
			\textsc{fmnist~/~mae} ~\,$\downarrow$ & $1.48 \pm 0.91$ & $1.42 \pm 0.91$ & $\maxf{1.39 \pm 0.89}$ & $1.11 \pm 0.02$ & $1.02 \pm 0.03$ & $\maxf{0.98 \pm 0.02}$ \\
			\textsc{fmnist~/~nlpd} $\downarrow$& $2.76 \pm 1.33$ & $2.71 \pm 1.31$ & $\maxf{2.65 \pm 1.23}$ & $2.16 \pm 0.18$ & $2.07 \pm 0.19$ & $\maxf{2.04 \pm 0.12}$ \\
			&&&&&&\\
			\textsc{cifar10~/~rmse} $\downarrow$& $2.66 \pm 1.08$ & $2.55 \pm 1.06$ & $\maxf{2.55 \pm 1.03}$ & $2.74 \pm 1.07$ & $2.64 \pm 1.08$ & $\maxf{2.57 \pm 1.02}$ \\
			\textsc{cifar10~/~mae} ~\,$\downarrow$& $1.77 \pm 1.06$ & $1.69 \pm 1.06$ & $\maxf{1.69 \pm 1.02}$ & $1.84 \pm 1.03$ & $1.76 \pm 1.05$ & $\maxf{1.71 \pm 1.03}$ \\
			\textsc{cifar10~/~nlpd} $\downarrow$& $3.20 \pm 1.55$ & $\maxf{3.07 \pm 1.44}$ & $3.32 \pm 1.89$ & $3.24 \pm 1.53$ & $3.14 \pm 1.53$ & $\maxf{3.06 \pm 1.45}$ \\
			\bottomrule
		\end{tabular}
		\label{tab:metrics}
	}
	{\scriptsize All metrics are ($\times 10^{-1}$).}
	\vspace*{-\baselineskip}
\end{table}

\subsection{Evaluation metrics}
In this section, we are interested in the evaluation of the GP decoder with the \emph{standard} error metrics used for GPs. In Tab.\ \ref{tab:metrics} we provide \textsc{rmse}, \textsc{mae} and \textsc{nlpd} for the three datasets considered in this experiment. Interestingly, the performance usually improves with larger active set sizes $A$, as the SAS model captures the underlying correlation of datapoints better and this enhances the approximation. This happens for both deterministic and Bayesian cases. We observe a similar trend for the \textsc{nlpd} metric which is generally better for larger active set sizes $A$. This can also be noticed in Fig.\ \ref{fig:sas_AS_curves}, where loss curves have a better convergence for the lowest $A$.

\textbf{Classification accuracy.}~~We are interested in evaluating the structure of the representation. For this purpose, we trained a (one) \emph{nearest neighbour} classifier on the encoded, two-dimensional latent variables and we tested the accuracy using encoded test data. Tab.\ \ref{tab:nn} shows the mean and standard deviations of test the accuracy.

\begin{figure}[ht!]
	\centering
	\includegraphics[width=\textwidth]{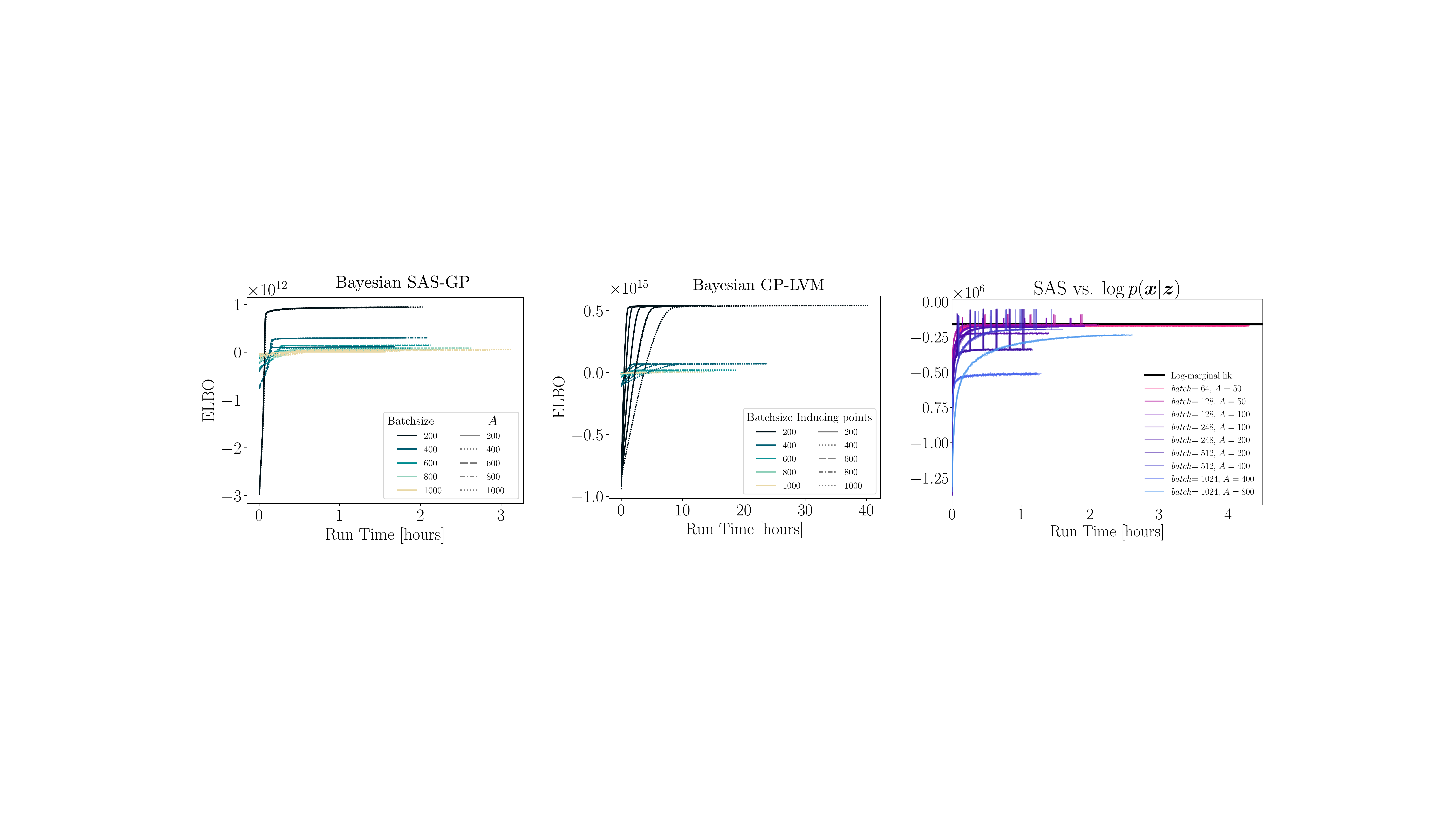}
	\caption{(\textbf{Left-center}) ELBO curves for Bayesian SAS-GP and Bayesian GPLVM for different batch-sizes and active set points. (\textbf{Right}) SAS loss function compared with the exact log-marginal likelihood computed. In all curves, $N$$=$$40,000$ samples of \textsc{mnist} were considered and five different initializations per \emph{batch} and $A$ setup.}
	\label{fig:loglik}
\end{figure}
\begin{table}[h]
	\caption{Classification accuracy ($\uparrow$) on 2-dim.\ latent space $\Zcal$.}
	\label{tab:nn}
	\centering
	\begin{tabular}{llll}
		\toprule
		\textsc{model}     & \textsc{mnist}    & \textsc{fmnist}\\
		\midrule
		\textsc{Bayesian sas-gp dec.} (\textbf{ours})  & $\maxf{0.63 \pm 0.022}$ &  $\maxf{0.63 \pm 0.020}$      \\
		\textsc{Bayesian gp-lvm}     &  $0.18 \pm 0.033$    &  $0.24 \pm 0.043$  \\
		\textsc{vae}  & $0.54 \pm 0.026$     &  $0.58\pm0.008$  \\
		\bottomrule
	\end{tabular}
\end{table}

\textbf{Runtime and convergence of SAS approximation.}~~The loss curves obtained for the (sparse) Bayesian GP-LVM model and the Bayesian SAS-GP decoder are shown in Fig.\ \ref{fig:loglik}. We observe that the two methods scale similarly, but that SAS is faster by a notable constant within stochastic optimization. Additionally, we observe the performance of the SAS approximation to closely fit the exact log-marginal likelihood $\log p(\xc|\zc)$. For the plot in Fig.\ \ref{fig:loglik}, we computed the exact probability for a subset of \textsc{mnist} $N$$=$$40,000$ samples.

\section{Conclusion}
\label{sec:conclusion}
State-of-the-art representation learning is generally based on neural networks, as this allows for scaling to large datasets. However, often we want reliable uncertainty estimates from the model and we can achieve these with Gaussian process decoders if we can scale them sufficiently. We have reviewed the main difficulties to obtain decent performances with GP-LVM approaches when applied to large-scale learning, even with inducing variables. Revisiting active set approximations, we considered a stochastic viewpoint to approximate the marginal likelihood while simultaneously keeping the model marginalized. We formulated our \emph{stochastic active sets (SAS)} approach for both deterministic and Bayesian versions of GP decoders. We found that our approach works well with amortization, such that a neural network encoder approximately inverts the GP decoder. While amortization also helps when using inducing points, we found the combination with SAS to be particularly efficient and robust. 

Empirically, we illustrated the advantage of our method first on image-based observations, where our approach learns better representations using fewer computational resources compared to inducing point methods. We further demonstrated that our approach easily scales to nearly $10^6$ observations. In this experiment, we found that the learnt representations are qualitatively on par with those attained by a comparable autoencoder. This is an important finding as, beyond small datasets, GP decoders generally recovers less useful representations compared with models based on neural networks. From this result, we speculate that improvements in \emph{training} might be enough to get state-of-the-art representations with GP decoders.

\paragraph{Additional benefits.} Besides the empirical benefits demonstrated by SAS in the previous section, we have also observed other practical benefits worth reporting. \emph{First}, we have observed that SAS easily runs in 32-bit numerical precision, unlike inducing point methods that generally require 64-bits of precision (when reporting running times we consistently used 64 bits). Similarly, the \emph{jitter} usually added to the Cholesky factorization is of less importance in SAS. \emph{Second}, we note that our implementation is surprisingly free of additional \emph{tricks} and no numerical heuristics were needed to realize a reliable implementation.

\paragraph{Limitations and future work.}

Stochastic active sets rely on the Gaussian likelihood, and this is perhaps the strongest limitation. This works well for continuous data, but many data sources are inherently discrete and this requires a suitable likelihood, e.g. the discretized mixture of logistics \citep{salimans_pixelcnn_2017}. Having more powerful likelihoods would surely improve the GP decoders, but this requires realization of further developments using SAS.

Future work will focus on applying the SAS approach in the supervised setting as well, and building SAS-like methods for discrete data. Other possible directions include extending the decoder with deep kernels \citep{wilson2016deep} to capture more features in the data and applying convolutional GPs \citep{van2017convolutional} which are more suited to high-dimensional images. 

\section*{Acknowledgements}

The authors want to thank Simon Bartels for the fruitful discussions during the early stages of our investigation. This work was supported by research grants (15334, 42062) from VILLUM FONDEN. This project has also received funding from the European Research Council (ERC) under the European Union's Horizon 2020 research and innovation programme (grant agreement 757360). This work was funded in part by the Novo Nordisk Foundation through the Center for Basic Machine Learning Research in Life Science (NNF20OC0062606). This work was further supported by the Pioneer Centre for AI, DNRF grant number P1.

\small
\bibliography{icml2022}

\newpage
\appendix

\renewcommand{\thesection}{\Alph{section}}

In this appendix, we provide additional details about stochastic active sets (SAS) as an approximation of the log-marginal likelihood for GP decoders, widely known as the \emph{Gaussian process latent variable model} (GP-LVM). We remark that SAS revisits \emph{active sets}, a sparse approximation predominantly used before the seminal work of \citet{snelson2006sparse}, in combination with stochastic optimization. The code for experiments is also included, and details on the data and initial setup of hyperparameters are included at the end of this appendix.

\section{Detailed derivation of Stochastic Active Sets}

The construction of SAS approximations for the log-marginal likelihood $\log p(\xc|\zc)$, builds on the connection between the \emph{evidence} and \emph{cross-validation} (CV) \citep{fong2020marginal}. The equivalence between \emph{leave-R-out} CV and the log-marginal likelihood is established by the use of predictive posterior scores, such that
\begin{equation}
    \mathcal{S}_{\text{CV}}(\xc|R) = \frac{1}{\mathcal{C}}\sum^{\mathcal{C}}_{p=1}\frac{1}{R}\sum_{n\in \Rcal_p} \log p(\xc_{n}|\xc_{\Acal_p},\zc) = \frac{1}{R}\Ebb_{\Acal_p}\left[\sum_{n\in \Rcal_p} \log p(\xc_{n}|\xc_{\Acal_p},\zc)\right],
    \label{eq:app_cv_score}
\end{equation}
where $\Acal_p$ denotes the \emph{active set} indices of the training data, such that $\Acal_p \subset \{1,2,\dots, N\}$ and $\Rcal_p = \{1,2,\dots, N\} \setminus \Acal_p$ are the remaining hold-out samples. The subscript $p\in \mathcal{C}$ denotes the permutation and we average over all $\mathcal{C} = \binom{N}{R}$ possible hold-out sets. We use use $R$ to indicate the size of the hold-out set $\Rcal_p$ and let $A=|\Acal_p| = N-R$. In particular, one might obtain the log-marginal likelihood in a cumulative manner by summing the scores $\mathcal{S}_{\text{CV}}(\xc|R)$ in Eq.\ \eqref{eq:app_cv_score} over all possible lengths of $R$,
\begin{equation}
    \log p(\xc|\zc) = \sum^{N}_{r=1}\mathcal{S}_{\text{CV}}(\xc|r),
    \label{eq:cv_log_marginal}
\end{equation}
which is the main result presented in \citet{fong2020marginal}. Notice that Eq.\ \eqref{eq:cv_log_marginal} has a similar computational cost as the exact calculus of $\log p(\xc|\zc)$ for GP decoders, since for small values of $r$, i.e.\ $r=1,2,3\dots$, we need to invert large covariance matrices $\bm{K}_{\Acal\Acal}$, where $A \rightarrow N$. Here, we drop the permutation subscript $p$ in $\Acal$ to avoid cluttered notation. Alternatively, we use the property that Eq.\ \eqref{eq:cv_log_marginal} can be factorised as

\begin{equation}
    \log p(\xc|\zc) = \mathcal{S}_{\text{CCV}}(\xc|R) + \mathcal{S}_{\text{PCV}}(\xc|R),
    \label{eq:cv_factorization}
\end{equation}
where $\mathcal{S}_{\text{CCV}}(\xc|R)$ is the \emph{cumulative} CV score and $\mathcal{S}_{\text{PCV}}(\xc|R)$ is defined as the \emph{preparatory} CV. Additionally, Eq.\ \eqref{eq:cv_factorization} holds for every size of the hold-out data $R \in [1,2,\cdots, N]$. This factorisation is of interest for us due to
\begin{equation}
    \mathcal{S}_{\text{PCV}}(\xc|R) = \sum^{N}_{r=R+1}\mathcal{S}_{\text{CV}}(\xc|r) = \frac{1}{\mathcal{C}}\sum^{\mathcal{C}}_{p=1}\log p(\xc_{\Acal_p}|\zc_{\Acal_p}),
    \label{eq:cv_pcv}
\end{equation}
where the r.h.s.\ term is equivalent to $\mathcal{S}_{\text{PCV}}(\xc|R) = \Ebb_{\Acal_p}[\log p(\xc_{\Acal_p}|\zc_{\Acal_p})]$. We remark that the computational cost of Eq.\ \eqref{eq:cv_pcv} is \emph{cheaper} than $\mathcal{S}_{\text{CCV}}(\xc|R)$ when the choice of $R$ is sufficiently large. Additionally, the \emph{cumulative} CV is defined as
\begin{equation}
    \mathcal{S}_{\text{CCV}}(\xc|R) = \sum^{R}_{r=1}\mathcal{S}_{\text{CV}}(\xc|r) = \sum^{R}_{r=1}\frac{1}{\mathcal{C}_r}\sum^{\mathcal{C}_r}_{p=1}\frac{1}{r}\sum_{n\in \Rcal_p} \log p(\xc_{n}|\xc_{\Acal_p},\zc),
    \label{eq:cv_ccv}
\end{equation}
where $\mathcal{C}_r = \binom{N}{r}$ are all the possible hold-out set for every value of $r$ considered. We remark the convenience of Eq.\ \eqref{eq:cv_ccv} for stochastic optimization as it includes predictive posterior probabilities $p(\xc_{n}|\xc_{\Acal_p},\zc)$ which emerge from the factorization of hold-$R$-out CV. This expression can be also rewritten as
\begin{equation}
    \mathcal{S}_{\text{CCV}}(\xc|R) =  \sum^{R}_{r=1}\frac{1}{r} \Ebb_{\Acal_p}\left[\sum_{n\in \Rcal_p} \log p(\xc_{n}|\xc_{\Acal_p},\zc)\right].
    \label{eq:cv_ccv_alt}
\end{equation}
The SAS approximation stochastically estimates both $\mathcal{S}_{\text{CCV}}(\xc|R)$ and $\mathcal{S}_{\text{PCV}}(\xc|R)$, with particular attention to the CCV score, which for large values of $R$ induces the largest computational cost. 

\section{Experiments, Algorithms and Metrics}

The code for the experiments is written in Python 3.7 and uses the Pytorch syntax for the automatic
differentiation of the GP models. It can be found in the repository \texttt{https://github.com/pmorenoz/SASGP}, where we also use the library Pyro for some baselines. In this
section, we provide a detailed description of the experiments and the data used, the initialization
of both latent variables $\zc$, the parameters of the \emph{amortization} network and hyperparameters $\bm{\theta}$. The training algorithms are provided in the main manuscript for both the \emph{deterministic} and Bayesian
approaches to the GP decoder. The performance metrics included in the main manuscript, e.g. the negative
log-predictive density (NLPD), the root mean square error (RMSE) and the mean absolute error (MAE).

\subsection{Detailed description and initialization}

All the models have matching encoding network architecture: three linear, fully connected layers with ReLU activation functions. The first two layers have sizes of 512 and 256 hidden units and the network encodes to two dimensions. The variances of the latent variables $\zc$ were encoded in a different way. In the VAE model, the variance was obtained by inputting the latent means to a soft-plus layer and the Bayesian SAS-GP and the Bayesian GP-LVM had separate network (with similar architecture) encoding the variance. The decoder in the VAE was built by a linear, fully connected layer with 400 hidden units, a softplus function and a sigmoid mapping. The GP-LVM baselines are implemented in Pyro \citep{bingham2019pyro}. Our implementation of the variational autoencoder is based on the official Pyro tutorial for VAEs. Importantly, the VAE and the SAS-GP implementations use standard \emph{data loaders} whereas the Pyro code must keep all the data in memory. This has limited the scaling possibilities of experiments with baselines. 

In most of our experiments, we use the \emph{vanilla} RBF kernel, where we initially set the \emph{amplitude} hyperparameter to $\sigma^2_a = 0.5$, the \emph{lengthscale} to $\ell = 0.1$ and the likelihood noise variance to $\sigma^2_n = 0.5$. The initial location of latent variables $\zc$ is subject to the initialization of the amortization networks, which is set up in a standard manner using the Pytorch \texttt{nn} module. Learning rates are set in the range $[10^{-4}, 10^{-2}]$ and the maximum number of epochs considered is 300.

\subsection{Datasets}

Our experiments make use of three well-known datasets: \textsc{mnist} \citep{lecun1998gradient}, \textsc{fmnist} \citep{xiao2017fashion} and \textsc{cifar10} \citep{krizhevsky2009learning}. All of them are downloaded from the \texttt{torchvision} repository included in the Pytorch library (\url{https://pytorch.org/vision/stable/datasets.html}). These particular datasets are not subject to use constraints or they include licenses which allow their use for research purposes.

\subsection{Error metrics}
Having defined the test dataset as $\xc_{*} = \{\xc_{n}\}^{N_*}_{n=1}$, we use the following error metrics to test the performance of the SAS approximation for GP decoders
\begin{align}
    \textsc{rmse}(\xc_{*}) &= \sqrt{\frac{1}{N_*}\sum^{N_*}_{n=1}(\xc_n - \bm{\mu}^{*}_{n} )^2},\\
    \textsc{mae}(\xc_{*}) &= \frac{1}{N_*}\sum^{N_*}_{n=1}\left|\xc_n - \bm{\mu}^{*}_{n} \right|, \\
    \textsc{nlpd}(\xc_{*}) &= \frac{1}{2}\log(2\pi) + \frac{1}{2N_*} \sum^{N_*}_{n=1} \left[ \log \bm{v}^{*}_{n} + \frac{(\xc_n - \bm{\mu}^{*}_{n} )^2}{\bm{v}^{*}_{n}}\right],
\end{align}
where $\bm{\mu}^{*}_{n}$ and $\bm{v}^{*}_{n}$ are the predictive mean and variance per $n$th test sample, respectively.

\subsection{Additional experiment with latent spaces of dimension larger than two}

We re-computed the experiments used for Table 2 using latent spaces of dimension \emph{three} and \emph{four}. The main outcome from these experiments is that training is as stable as in the former cases with two dimensions in the latent space. In general, we observed a similar performance as in the rest of experiments included in the main manuscript. So we remark that there is no limitation in our framework to accept $\text{dim}(\mathcal{Z}) > 2$.

\begin{figure}[ht!]
	\centering
	\includegraphics[width=0.5\textwidth]{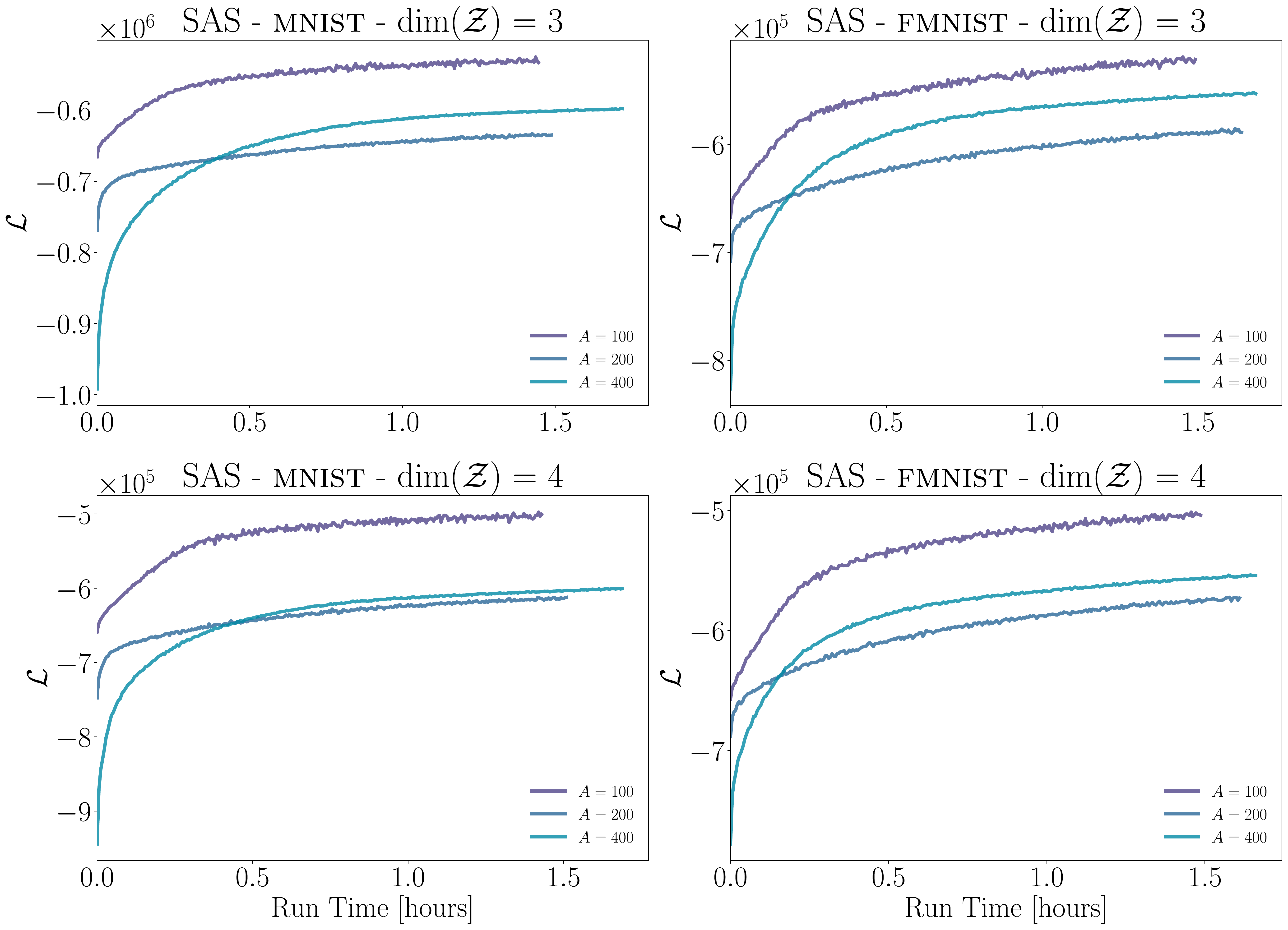}
	\caption{Training curves for different active set sizes $A$ and dimensionalities of the latent spaces. Each row contains the results for \textsc{mnist} and \textsc{fmnist}, respectively.}
\end{figure}

\subsection{Ablation study}

We did additional experiments as an ablation study based on Eq. 6. In particular, we ran the SAS model for $A=\{100,200,400\}$ using the \textsc{mnist} and \textsc{fmnist} datasets. The first ablation experiment shown in Figure 2 corresponds to using only the \emph{second term} $\log p(\bm{x}_\mathcal{A}| \bm{z}_\mathcal{A})$ in Eq. 6. We can observe that the performance is not as good as in the results illustrated in the main manuscript. Alternatively, we also included another ablation experiment using the \emph{first term} of Eq. 6, which is shown in Figure 3. In this last case, the performance is not good and the structure in the latent space is only provided by the amortization net.

\begin{figure}[h!]
\centering
\includegraphics[width=0.75\textwidth]{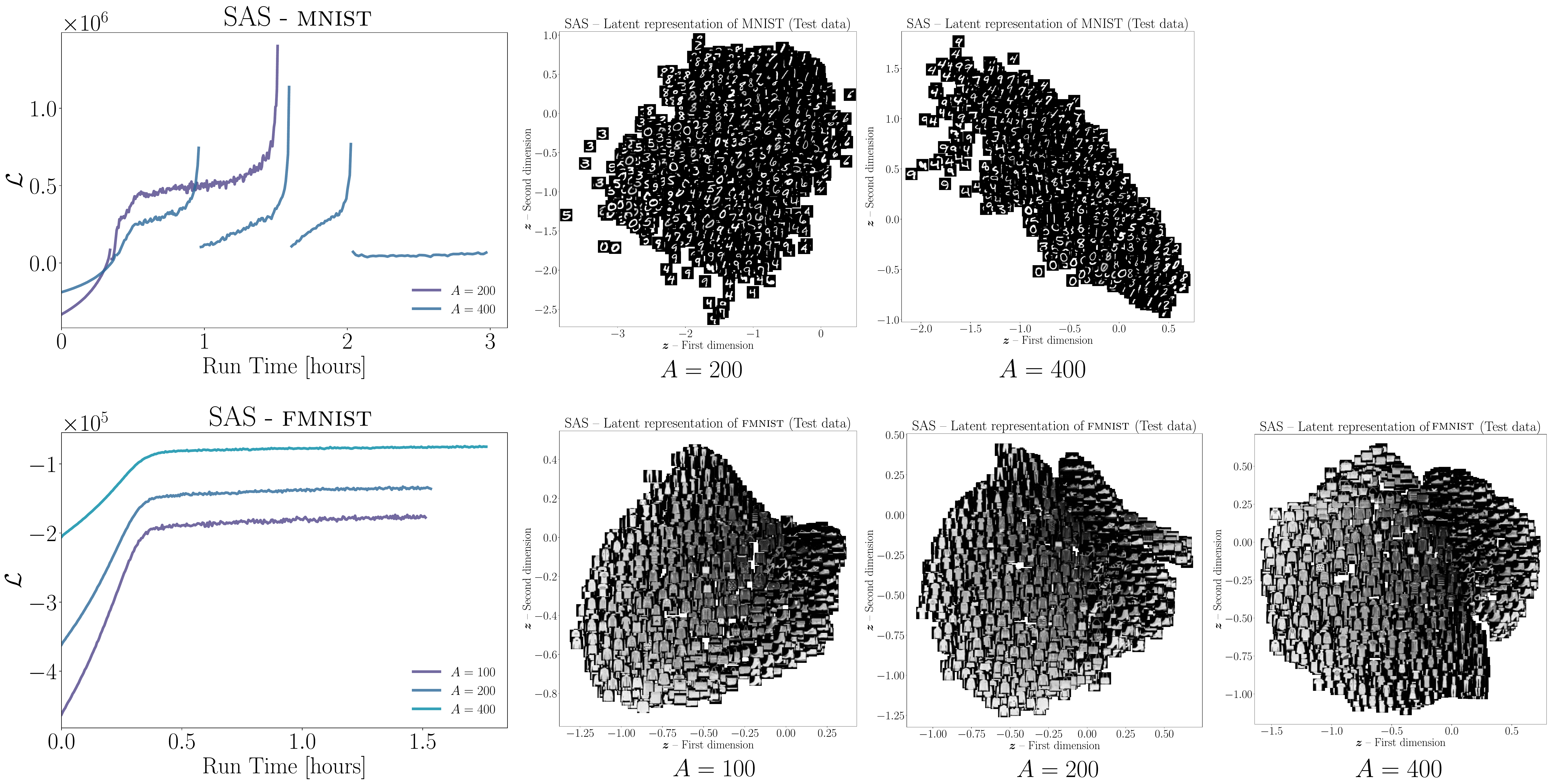}
\caption{Ablation study for Eq. 6.\ The approximation of the log-marginal likelihood is only computed with the firm term (factorisation). Curves are computed for $A=\{100,200,400\}$ and rows indicate the dataset \textsc{mnist} or \textsc{fmnist}.}
\end{figure}

\begin{figure}[h!]
	\centering
	\includegraphics[width=0.75\textwidth]{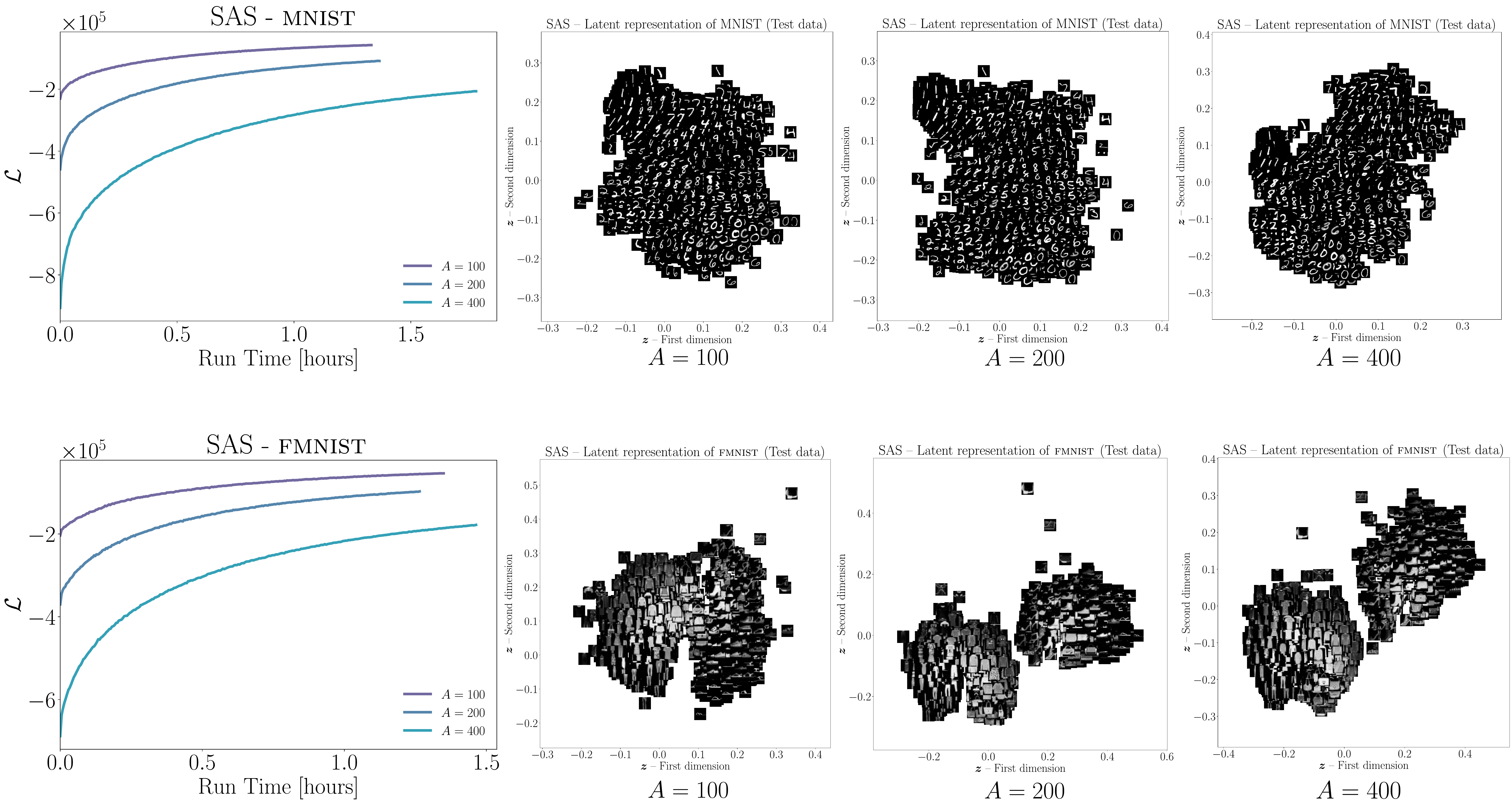}
	\caption{Ablation study for Eq. 6.\ The approximation of the log-marginal likelihood is only computed with the second term (full covariance). Curves are computed for $A=\{100,200,400\}$ and rows indicate the dataset \textsc{mnist} or \textsc{fmnist}.}
\end{figure}


\end{document}